%% file: 0_main.tex
\documentclass[letterpaper, 10 pt, conference]{ieeeconf}  % Comment this line out if you need a4paper

\IEEEoverridecommandlockouts                              % This command is only needed if 
\usepackage{graphicx}
\usepackage{float}
\usepackage{comment}
\usepackage{array}
\usepackage{soul}
\usepackage{stfloats}

\title{\LARGE \bf Sim2real gap is non-monotonic with robot complexity for morphology-in-the-loop flapping wing design}

\author{Kent Rosser$^{1+}$, Jia Kok$^{1}$, Javaan Chahl$^{2}$ and Josh Bongard$^{3}$
\thanks{*This work was supported by the DST Group Trusted Autonomous Systems Strategic Research Initiative and the Vermont Advanced Computing Core}% <-this % stops a space
\thanks{$^{1}$Defence Science and Technology Group, Australia}
\thanks{$^{2}$University of South Australia, Australia}
\thanks{$^{3}$University of Vermont, VT USA}
\thanks{$^{+}$kent.rosser@dst.defence.gov.au}
}
\begin{document}

% \title{Sim2real gap is non-monotonic with robot complexity for morphology-in-the-loop \RSSFIXME{flapping wing} design}

% avoiding spaces at the end of the author lines is not a problem with
% conference papers because we don't use \thanks or \IEEEmembership


% \author{
% % \authorblockN{
% % Kent Rosser$^{1,2,3}$,\hspace{4pt}
% % Jia Kok$^{1}$,\hspace{4pt}
% % Javaan Chahl$^2$,\hspace{4pt}
% % Josh Bongard$^3$
% % }

% % \authorblockA{$^1$\hspace{1pt}Defence Science and Technology Group, Australia}
% % \authorblockA{$^2$\hspace{1pt}Department of Engineering and Maths, University of South Australia}
% % \authorblockA{$^3$\hspace{1pt}Department of Computer Science, University of Vermont}

% % \authorblockA{Email: kent.rosser@dst.defence.gov.au}

% \IEEEauthorblockN{Kent Rosser}
% \IEEEauthorblockA{Defence Science and Technology Group, Australia\\
% kent.rosser@dst.defence.gov.au}\\
% \IEEEauthorblockN{Jia Kok}
% \IEEEauthorblockA{Defence Science and Technology Group, Australia\\
% jia.kok@dst.defence.gov.au}
% \IEEEauthorblockN{Javaan Chahl}
% \IEEEauthorblockA{Dept.~of Engineering and Math, University of South Australia\\
% javaan.chahl@unisa.edu.au}
% \IEEEauthorblockN{Josh Bongard}
% \IEEEauthorblockA{Dept.~of Comp.~Sci.\\University of Vermont\\
% josh.bongard@uvm.edu}

% }

\maketitle

%------------------------------------------------------------------------- 
% Teaser
\begin{figure*}[b]%
\centering
\includegraphics[width=\linewidth]{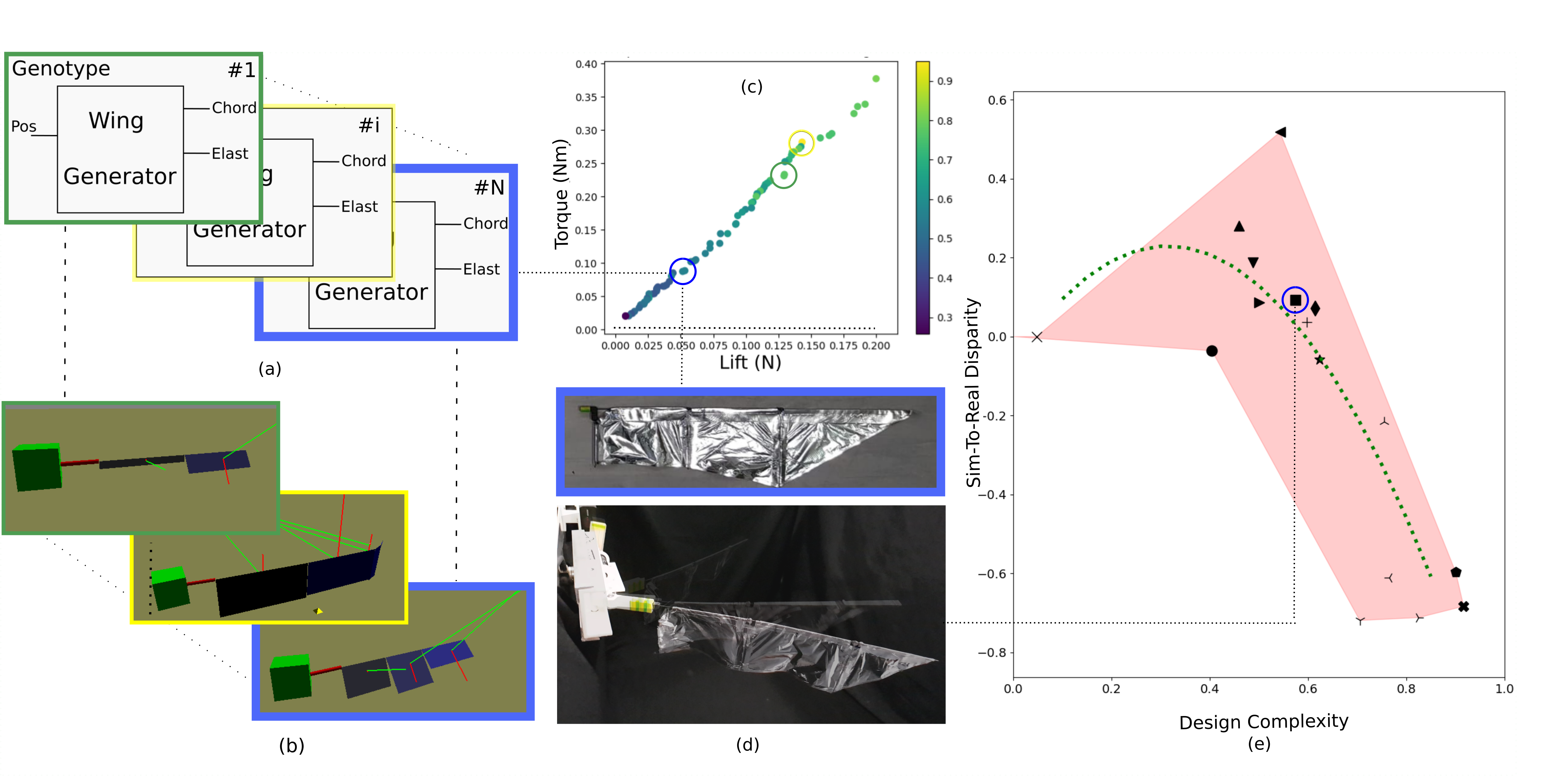}\\
\vspace{-15px}
\caption{Automatic design of flapping wing morphologies using simulation produces a set of high performance designs that are transferred to reality to measure the shape of the simulation to reality gap. (a) An evolved population of designs (genotypes); (b) designs are evaluated in simulation; (c) evolutionary algorithm produces a Pareto optimal set of designs; (d) selected individual is transferred to reality and evaluated; (e) plot of the simulation-to-reality gap defined by transferred designs.} 
%parameters which define a proposed solution to the problem that the genetic algorithm is trying to solve. 
%Using a parameterised phenotype tailored to both simulation and low cost manufacture, we use evolutionary algorithms to design a population of robotic flapping wing morphologies in simulation that maximise Lift while reducing actuator requirements.  We transfer a selection of high performance wings from simulation to the reality to measure the shape of the reality gap.} 
\label{fig:teaser}
\end{figure*}
%------------------------------------------------------------------------- 

\begin{abstract}

\input{1_abstract}

\end{abstract}

\vspace{5px}
\textbf{Keywords - morphology ;  simulation to reality ; evolution ; bio-inspired }

\input{2_intro}

\input{3_methods}

\input{4_results}

\input{5_discussion}

\bibliographystyle{plain}
\bibliography{refs4evo.bib}

\end{document}

%% file: 1_abstract.tex
%Morphology of a robot design is important to its ability to achieve a stated goal and therefore applying machine learning approaches that incorporate morphology in the design space provide scope for significant advantage.  Our study is set in a domain known to be reliant on morphology: flapping wing flight. We developed a parameterised morphology design space that draws features from biological examples and we apply automated design to produce a set of high performance robot morphologies in simulation. By performing sim2real transfer on a selection, for the first time we measure the shape of the reality gap for variations in design complexity.  We find for the flapping wing that the reality gap changes non-monotonically with  complexity, suggesting certain morphology types may narrow the gap more than others.

Morphology of a robot design is important to its ability to achieve a stated goal and therefore applying machine learning approaches that incorporate morphology in the design space can provide scope for significant advantage.  Our study is set in a domain known to be reliant on morphology: flapping wing flight. We developed a parameterised morphology design space that draws features from biological exemplars and apply automated design to produce a set of high performance robot morphologies in simulation. By performing sim2real transfer on a selection, for the first time we measure the shape of the reality gap for variations in design complexity.  
We found for the flapping wing that the reality gap changes non-monotonically with complexity, suggesting that certain morphology details narrow the gap more than others, and that such details could be identified and further optimised in a future end-to-end automated morphology design process.

%% file: 2_intro.tex
\section{Introduction }

In many applications, the design of a robot is intuitive, fast and achievable by human hand.  A fixed morphology robot platform design can then be produced and well characterised into a simulation model, allowing control policies to be developed that achieve a desired task in simulation and then subsequently transferred to reality \cite{jakobi1995noise,peng2018sim,koos2013transferability,mordatch2015ensemble,hwangbo2019learning}, a process known as sim2real transfer.  For some applications however, the intuitive design approach to morphology may not be available as the construction materials, actuation and the environmental forces being experienced by the robot may not be within the realm of traditional design experience \cite{rus2015design}. Without a guiding intuition, manual design can degrade into a systematic form of trial and error and applying machine learning approaches to morphology design becomes appropriate. 

Evolutionary algorithms are one such machine learning approach capable of producing a population of designs from which to select \cite{sims1994evolving,sims1992interactive}. Inspired by the ability of natural evolution to produce diverse and complex structures, evolutionary robotics has developed tools that facilitate  morphological search to produce robots with novel morphologies \cite{cheney2018scalable} that a human designer may not consider \cite{lehman2018surprising}. 
%and that traditional machine manufacture could not create \cite{guo2013additive}.  

Automated design of these robots in simulation is challenged by the ``reality gap"  problem \cite{jakobi1995noise}. The reality gap is the observed difference between the performance of a robot with high fitness in simulation and its real world equivalent, and remains a critical issue for robotics \cite{doncieux2015evolutionary,jakobi1995noise,lipson2000automatic,koos2013transferability}.  The ability to produce simulations that match reality for arbitrary behaviour and morphology exceeds currently available physics simulation packages, and when we consider the effect of material properties, non-homogeneous materials, stiffness, aerodynamics and friction (to name a few) it is easy to imagine a range of ways that robots designed in simulation can produce behaviours that will differ from reality.

% Introduced here \RSSFIXME{R1 - INTRODUCE intuitively LINEAR AS BASE ASSUMPTION - WHY?}

% Trimming May 2019 - this paper is not strongly focused on manufacturing  rather the reality gap. 
% 
%Morphology-in-the-loop design has also been coupled to additive manufacture \cite{lipson2000automatic} with an expectation that in the future 3D printers will be ubiquitous, cheap and capable of printing arbitrary shapes with arbitrary materials and high resolution. Although substantial progress has and continues to be made, the current reality is that 3D printers can print only a limited set of materials with material properties that vary depending on the layering during manufacture \cite{ngo2018additive}, and that the efficiency and response rate of conventionally designed and manufactured electro-mechanical motion is still well ahead of artificial muscle actuation \cite{miriyev2017soft}.

There is a growing literature reporting machine learning design methods that automatically achieve successful sim2real transfer without using human intuition to improve the simulation or undertake the design. Jakobi \cite{jakobi1997evolutionary} produced early work in sim2real transfer for fixed morphology robots by applying noise to the simulation of the components that would likely transfer poorly. More recently Hwangbo \textit{et al} \cite{hwangbo2019learning} showed that for a fixed morphology robot the reality gap  can be mapped into a deep neural network facilitating sim2real transfer.  Lipson \textit{et al} \cite{lipson2000automatic} reported the first example of sim2real design with both morphology and control.  Their work demonstrated the existence of the reality gap for morphological design, as only a small fraction of their realised robots performed to expectation.  Bongard \textit{et al} \cite{bongard2006resilient} successfully crossed the reality gap by developing a method for continuous morphology self-modelling while operating on the physical robot and co-evolving a set of controllers to suit in simulation.  Koos \textit{et al} \cite{koos2013transferability} produced the transferability method to automatically achieve sim2real transfer for controllers. That method undertakes sim2real transfer of selected designs, quantifies the difference between simulation and real performance and for any poor transfer, it probabilistically excludes the local region of the design space from further search.  While all of these examples have made contributions to the automatic design of robots, none of them have provided any machine learning methods that address reliable sim2real transfer when morphology is included in the design loop, nor provided any explanation as to why certain regions of the design space do not transfer well that can be exploited by an automated design process.

To begin to address these gaps when morphology  is not fixed (which we call morphology-in-the-loop design), we seek an understanding of 'shape' of the reality gap.  In variable morphology simulation, it is typical to model a design as a set of relevant finite elements such as rigid body components, voxels, splines etc \cite{lipson2000automatic, cheney2014evolved, collins2018towards} and accumulate the localised effects experienced by each element to produce an overall robot behaviour.  We know from FEA (Finite Element Analysis) tools used extensively in human directed computer aided design to deconstruct larger problems into smaller elements, that simulation accuracy is dependant on the number of elements and the overall size of a design \cite{liu2013effects}, both of which vary in a  morphology-in-the-loop design process. Using a finite element approach to morphology design adds morphological structure by adding one or more additional finite elements, so it is consistent to expect that simulation accuracy will relate to the total number of finite elements as well as the spacing of those elements across the size of the design. In this study, we use the term morphological simulation complexity to represent that combination of elements and size of a robot design within a specific simulation approach.  {Qualitatively, when a robot morphology with known simulation error is extended by adding another finite element representing additional structure, the initial simulation error is expected to be retained and the new element will incorporate its own model imperfections as an additional quantity of error, intuitively suggesting that simulation error would grow  monotonically with morphological simulation complexity.}

We explore this reality gap shape within the bio-inspired field of flapping wing flight.  Flapping wing aerodynamics remains a challenging unsolved problem \cite{shyy2010recent} and  customisation of morphology can be intuitively expected based on the specific objective to be optimised.  While only a few man-made fliers exist in this domain \cite{nakata2011aerodynamics}, natural fliers have solved the problem of staying airborne through evolution, and their morphology is known to be important and tailored to their environmental niche. Consider some examples of natural fliers. Dragonflies have four long, high aspect ratio wings that provide agility; butterflies have short low aspect ratio wings flapped at a lower rate allowing long flight times; and birds have wing geometries dependant on species that are tailored to speed, manoeuvrability or endurance. Beyond just wing shape, the distribution of mass, elasticity and actuation are also important.  For example, Additionally, Yin \textit{et al }\cite{yin2010effect} showed that mass of the wing is critical for flapping wing systems, and  Li \textit{et al} \cite{li2017wing} showed the substantial amount of angular twist ($60^o$) across the span of a dragonfly wing during maneuver, which can reasonably be expected to be enabling rather than detrimental to its performance.  Examples such as these suggest morphological design is crucial for gaining a niche in an environment, and equally that no control scheme can overcome limitations caused by a poor morphology.

Human designers often seek inspiration from biological examples that natural evolution has honed over aeons to achieve spectacular refinement and capability.  Since natural systems are evolved into their niches rather than ones that humans desire, rather than direct mimicry of the morphology of biological wings, we approached the challenge of morphological design by identifying salient characteristics of biological examples for use as building blocks in a machine learning approach that is able to automatically specify a morphological design.

This paper presents our sim2real morphology-in-the-loop design within our parameterised morphospace (the region spanned by all specifiable morphology design solutions) of a flapping wing aerodynamics problem.  We have created an evolvable design specification (genotype) that can be expressed to create a single structured flapping wing robot (phenotype) for analysis in a simulation engine or to form the specification for the manufacture and assessment in reality.  Our manufacturing process allows each wing design to be produced using readily available materials in a short time frame and at low cost, allowing this study to include substantially more morphology transfers than previous sim2real studies cited above.%  (see tables \ref{tab:wingMaterials} and \ref{tab:manuTimeBreakdown}).  

Here, for the first time, we measure the shape of the morphology design reality gap {as a function of morphological complexity}. For our flapping wing system the shape is {non-monotonic}, which has implications for machine learning approaches to simulator development and morphology-in-the-loop design, and sets future research questions relating to the generality of the shape and the factors that drive it.

%\RSSFIXME{R2 - The concept of morphospace is repeatedly mentioned without any attempt to properly quantify this space. If this is a parametric space defined by all possible axes of design research, a more principled approach is needed to identify the key variables that quantify this space and their relationship. If not, there is a strong risk of obtaining results that are not rigorous and not really systematic. For instance, how many variables quantify the space being spoken of in this manuscript? How are scaling / resolution issues dealt with? Are they continuous or discrete? The only constraint dealt with seems to be in the max lift produced but that does not quantify in any manner constraints on the underlying variables themselves. This to me calls in question the quality of results claimed.}

%% file: 3_methods.tex
\section{Materials and Methods } 

{This section describes the bio-inspired flapping wing morphospace used in this study,  From the introduction, morphological shape, size, stiffness and inertia are known to be important for natural fliers so we made these features available for selection within our morphospace including the simulator, real world fabrication and evolutionary design approaches. }
%Further details of these are available in the supplementary data.  

\subsection{Simulator}

%Flapping wing aerodynamics is a developing field of aerodynamics.  Because of the complexities, many studies in this domain use computational fluid dynamics (CFD) and wind tunnel testing (WTT) , however, as well as fluidic concerns, wing structure deflections under load complicate this further \cite{chimakurthi2009computational}. Although rarely reported, evaluation times for even simple instance in CFD are computationally expensive \cite{collins2018towards} and increase at least linearly when performed in sequence over many time steps, and WTT is constrained by equipment costs.

Within flapping wing research, many studies use computational fluid dynamics supported by wind tunnel testing (WTT) \cite{khurana2013bioinspired,nakata2011aerodynamics} for the analysis of a single wing design, however those tools are poorly suited to evaluation of large numbers of different designs due to computation time, finite element meshing variation and fluid structural interaction \cite{jones2013cfd}.  We are driven towards approaches relevant to automated design, and for that reason we hypothesise that CFD may not be required.  Instead we propose a lower fidelity but faster simulation process that allows a much wider exploration of the search space. {Provided a simulation provides a fitness landscape with a gradient towards high performance solutions in reality, it is suitable for automated design.}

\begin{figure}[t]
%   \centering
% %    \includegraphics[width=0.4\textwidth]{GenotypeExample.png}
%     \includegraphics[width=0.48\textwidth]{images/genotype.png}\\
%     (a)  
  \centering
(a)    \includegraphics[width=0.45\textwidth]{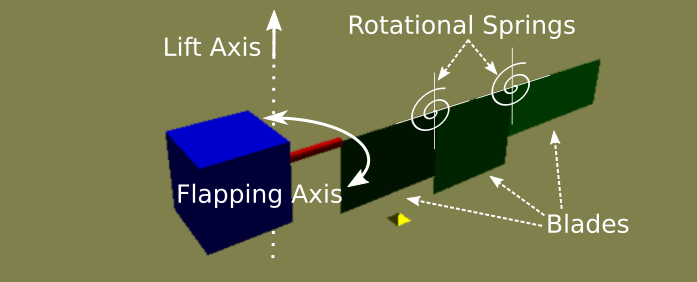}
%(b)    \includegraphics[width=0.45\textwidth]{images/NDF67MAP1template_BIGFONT.png}(b) fabrication mapping of genotype to physical wing
(b)     \includegraphics[width=0.45\textwidth]{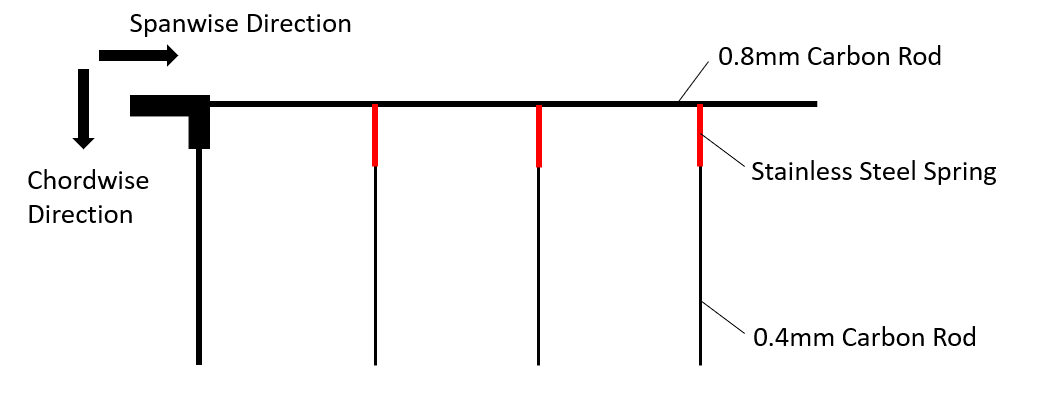}
(c)    \includegraphics[width=0.45\textwidth]{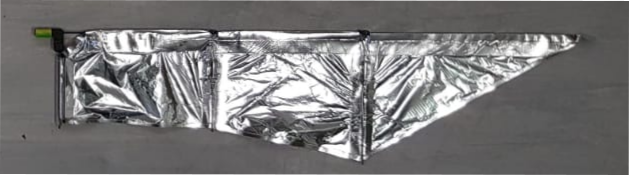}
(d)    \includegraphics[width=0.45\textwidth]{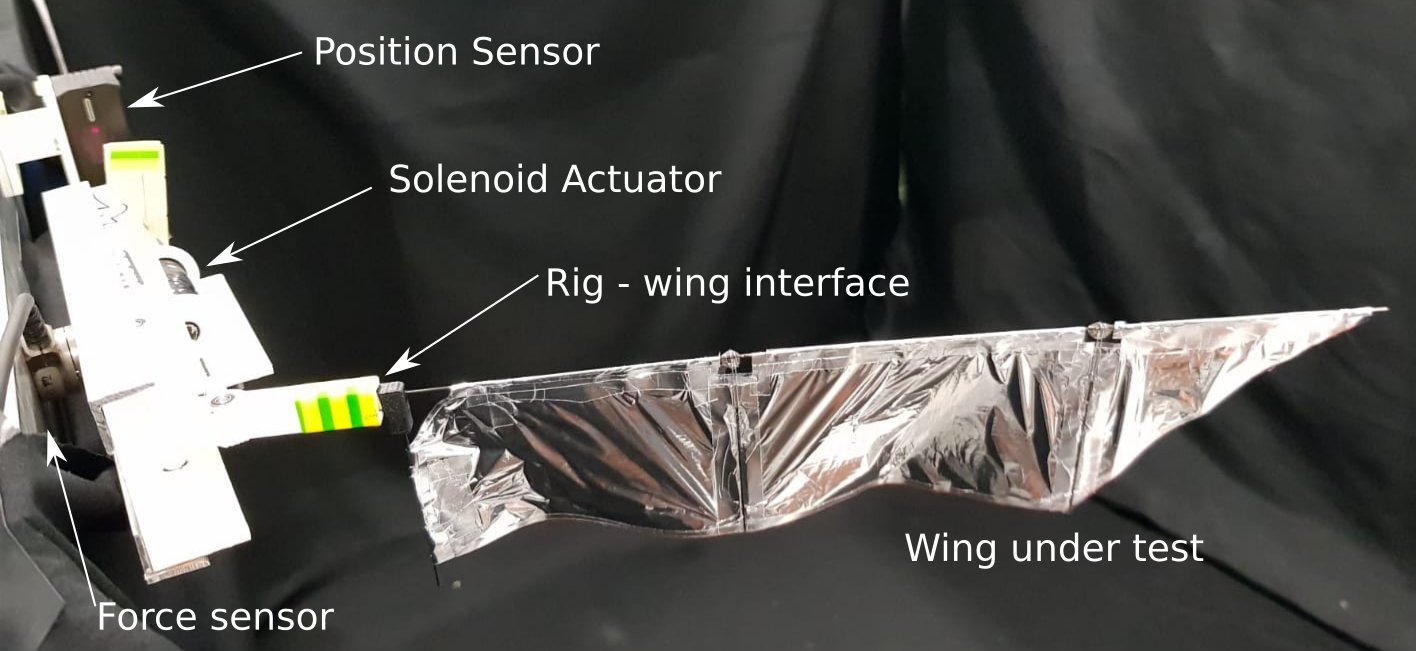}
  \caption{Robot wing {development pipeline}  (a) simulation phenotype defined by genotype for evaluation  (b) construction of elastic wing ribs (c) realised wing after manufacture (d) wing mounted to flapping wing test apparatus.}
  \label{fig:exampleWingSequence}
  \vspace{10px}
\end{figure}

For our flapping wing simulation, we modelled a flapping wing design as a set of spanwise aligned connected flat "blade" elements.  Each individual blade includes local geometry (span and chord) values that collectively determine the shape, span and inertia distribution of the entire wing.  The blades are sequentially connected by joints that allow  elastic deformation in both chordwise (twisting) and spanwise (bending) directions when the wing is under load.  Figure \ref{fig:exampleWingSequence}(a) shows a full wing constructed in the simulator.

Each blade embodied the Sane \textit{et al} quasi-static model for flapping wing aerodynamic forces and torques \cite{sane2002aerodynamic}.  That model solves time domain forces  as the sum of translational and rotational effects.  It excludes wake capture, which is a poorly described force in the literature related to the interaction of the wing with the disturbance it created in a previous stroke.

% Added brief summary how lift is induced. \RSSFIXME{R3 CLARIFY WHY 1DOF and can lift be produced? } 

The simulation was encapsulated in the PYROSIM robotics simulator \cite{kriegman2017simulating} which was extended to compute quasi-static forces for each blade within a simulation, while the underlying physics simulator resolved the inertial and elastic response of the wing.  The wing is connected by a wing root (red cylinder) to a rigid base (blue cube).  Rotational actuation of our flapping wing simulation was applied at the root of the wing around the z-axis, with angle controlled in accordance with a provided control waveform.  Figure \ref{fig:teaser}(b) shows simulation of three differing individuals and visualises the instantaneous quasi-static normal and axial forces (green and red lines) for each blade. 

Actuation causes interaction between the environment and the wing and therefore morphology dependent amounts of lift and drag as the wing translates and deforms. The base block in the simulator measures the force and torque experienced during actuation cycles to determine overall lift (vertical force) experienced and drive forces applied.

% reworded - \RSSFIXME{R3 UPDATE WORD - timeliness -> }

% In contrast to the significant computational demands of CFD, our quasi-static aerodynamics simulation for an arbitrary wing controlled by a arbitrary flapping waveform is able to reliably simulate two seconds of arbitrary wing activity within 100ms wall clock time on a single core of a desktop computer. This execution time of the simulator is well suited to evolutionary design research and development.

%In summary, we have developed a low computation cost simulation of an elastically deformable flapping wing robot that interacts with its local environment to behave in accordance with a standard quasistatic model.

\subsection{Manufacture and Test}

%\textbf{** Describe that wing constrution is defined by the genotypy but that our prcess has foccused on a 1-1 correlation of finite elements -> artefact in phys.??  If it was arbitrary then gap could be anything!}

To support sim2real transfer, we developed a cost-effective manufacturing method that produces a wing that allows controlled variation of span, shape, elasticity and inertia.   

%Contrary to fixed and rotary wing airframe designs, a bio-inspired wing system requires wings that are flexible.  It is the proper coupling of the aerodynamic and elastic forces that produce better performing wings.

% Trimmed May 2019 - inertia is repetive to next para.
%Wing design and manufacture in any flying body is critical to its performance, and limiting overall system weight is often a primary design driver. In flapping wing systems, it has been shown that the weight of the wing is even more critical due to the additional need to constantly accelerate and decelerate the wing \cite{yin2010effect}. Reducing the inertia of the wing reduces the power requirements of the system and expands the boundaries of the feasible morphospace. 
 
From the wing root, a carbon fibre rod formed the main spar of the wing and its length set the span of the physical wing.  One of the morphological parameters in the morphospace is the elasticity along the wing, which impacted on the amount of aero-elastic deformation that the wing experiences when under load. It is difficult and costly to accurately produce a smooth elasticity profile along the wing, so we discretize the elasticity by attaching ribs to the spar of the wing  oriented in the chordwise direction, connected with spring elements as illustrated in Figure \ref{fig:exampleWingSequence}(c). The spring elements used were stainless steel music wire of varying diameters creating selectable levels of elasticity. Spring elements were isolated to a 15mm section of the rib attached to the spar, while the remainder of the rib element was formed by a rigid carbon rod stiffener of dimension such that the rib length is the same as the wing shape geometry at that position on the wing. Using this, we obtain a definable shape, elasticity and inertia profile along the span without adding significant manufacturing complexity.  The aerodynamic skin was aluminised Mylar film, which was selected for its robustness.
A completed wing approximately weighed 1 gram, costs US\$1 in consumables and required approximately 2 hours of assembly time (excluding adhesive cure time).

To assess real world performance, a flapping wing test rig was developed and is depicted with a mounted wing in Figure \ref{fig:exampleWingSequence}(d).  Our test rig drives the wing repeatedly through a single flapping axis using a linear solenoid actuator with magnetic plungers similar to the design demonstrated by Kok \textit{et al}.~\cite{kok2016design}. Position feedback was provided by a laser displacement sensor (Keyence IL100) that measured the position of the magnetic plunger which defined angular position of the wing. An Arduino Due \cite{arduino} measured the position feedback and commanded the actuator such that real time angular position of wing stroke could be accurately controlled. The actuator and wing assembly was mounted on an ATI Nano17 Force/Torque transducer such that forces generated by the actuated wing were recorded by a National Instruments data acquisition system \cite{niusb6003} at a sampling frequency of 100kHz. The flapping test rig was capable of producing controlled oscillation beyond 10Hz dependant on the specific wing attached over a angular range of $\pm40^o$. This study set a fixed sinusoidal pattern at 5Hz to facilitate arbitrary  morphology-in-the-loop design without risking damage to either the wing or apparatus.

\subsection{Evolutionary design}

To conduct automated design using evolution, we defined a genotype structure that could describe designs that span of the morphospace.   Figure \ref{fig:teaser}(a) shows an image of a population of individuals that are each able to specify a unique wing morphology by evaluating their own wing generator at a set of evolved positions along the wing span.  The result is a wing design that includes variations in shape, elasticity and inertia along its span which matches the bio-inspired design space made available in simulation and manufacture.

We placed the development of a population of these genotypes under evolutionary control in simulation.  The algorithm used for this study was the Non-dominated Sorting Genetic Algorithm II (NSGA-II) \cite{deb2002fast} evolutionary optimisation method.  Multi-objective optimisation is commonly used to maintain a diverse set of high performance designs within an evolving population.  In this case, we searched for designs that maximised Lift produced while minimising drive power and torque.  In optimising across more than one dimension, rather than determining a single best morphology, the search tool produces a set of individuals known as a non-dominated front (NDF). Within a NDF, each individual represents a single design that is superior to all the others in the population on at least one objective.  Figure \ref{fig:teaser}(c) shows one example NDF produced using this method.

\subsection{Experiment}

% Trimmed for CoRL - repetitive
%In the evolutionary robotics design field, there has been limited success in the transfer of individual solutions formed by simulation to real world systems, but only superficial discussion of why some phenotypes transfer and others do not, particularly in morphology-in-the-loop design.  In this experiment, we sought to form a better understanding of the simulation to reality gap than has previously been documented.

%Using the parameterised phenotype simulation and manufacture concept introduced, we measured the reality gap for a set of individual genotypes by comparing their performance results in simulation with those measured when produced in reality.

For our experiment we were interested in the variation between simulation and reality of high performance solutions as we vary the {structure of the design specification in our parameterised design space}.  To focus on morphological effects, we constrained the control behaviour in both simulation and reality to be a fixed 5Hz sinusoid, while we varied the wing design to produce efficient lift in simulation.  We transferred high performing designs to reality and compared the lift as measured from the flapping wing test rig to determine transferability.  

In their study, Koos \textit{et al} \cite{koos2013transferability} defined the Simulation To Reality ($STR$) disparity measure to assess transferability of controllers as the difference between the phenotypic performance measures in simulation and reality. Their definition of $STR$ included an built in bias which matched their observation that only reduced performances were observed following simulation to reality transfer - i.e. the controller always performed worse in reality than simulation. We propose a refined definition to allow for the possibility that reality may outperform simulation in morphological design.  

%% REmoved following feedback in CoRL which showed that in N-dimensions diverged sim2real behaviour can cancel the STR metric to 0 which is not intended.  It works for our 1D case though so remove general form and for the specific approach.
%For any robot $m$ in morphospace $M$, the $STR(m)$ is given by Equation \ref{eqn:str_defn}.
%
%\begin{equation}
%STR(m) = \frac{1}{T}\sum_{t=1}^{T}{\frac{R_t(m) - S_t(m) }{S_t^{MAX}}}
%\label{eqn:str_defn}
%\end{equation}
%
%where $S_t(m)$ and $R_t(m)$ represents the $t$th performance metrics of the $m$th design in simulation and reality.  $S_t^{MAX}$ is the maximum value of the $t$th metric for all transferred solutions.  Using this definition, an $STR$ value near zero represents good simulation to reality transfer as the discrepancy is small compared to predicted.  Positive $STR$ implies reality has outperformed simulation and negative $STR$ shows a degradation after the transfer.

For our flapping wing experiment, the phenotypic performance metric we use is lift. For any robot design $m$, we calculates $STR$ as per Equation \ref{eqn:str_defn_fw}, where ${L_{S}}$ and ${L_{R}}$ are the mean inter-cycle lift per flapping cycle averaged over 5 flapping cycles for simulation and reality respectively. ${L}^{max}$ is the maximum value of $L_{S}$ of all wing designs considered.

\begin{equation}
STR(m) = \frac{{L}_{R}(m) - {L}_{S}(m) } {{L^{max}}}
\label{eqn:str_defn_fw}
\end{equation}

% Discuss with Josh - is D_M more of a simulation finite element thing?
%  \RSSFIXME{R2 - Clarify Distance metric - including equation.  **It was in there!**}
% \RSSFIXME{R3 EXPLAIN HOW WE GET $D_M(_m)$ equation } 

%We are assessing the variation in $STR$ with respect to \RSSFIXME{morphological simulation complexity}.  From the introduction, we identified that the number of elements and their location across the size of the design were important to simulation accuracy.  
%We define a quantitative measure,  $C_{MS}$, for any wing design $m$ in this study using Equation \ref{eqn:morphSimComplex_defn_}. 
%\begin{equation}
%C_{MS}(m) = \frac{1}{N}\sum^N_{n=1}\frac{M_i(m)}{M_i^{max}}
%\label{eqn:morphSimComplex_defn}
%\end{equation}
%\RSSFIXME{Where $M_i(m)$ is the $i$th phenotypic metric of the finite elements structure within the morphology simulation for design $m$} and is normalised by the maximum value of that metric used across all designs transferred within the experiment, $M_i^{max}$. 

We are assessing the variation in $STR$ with respect to {morphological simulation complexity}, $C_{MS}$.  From the introduction, we identified that the number of  finite elements and their location across the size of the design were important to simulation accuracy, so we define this measure for a design $m$ in our flapping wing morphospace as per Equation \ref{eqn:morphSimComplex_defn_fw}.

\begin{equation}
C_{MS}(m) = \frac{1}{2}\Big(\frac{B(m)}{B^{max}} + \frac{S(m)}{S^{max}}\Big)
\label{eqn:morphSimComplex_defn_fw}
\end{equation}

where $B(m)$ is the number of blade finite elements in the simulation and $S(m)$ is the span wing design. $B^{MAX}$ and $S^{MAX}$ are the maximum value of those measures across all designs.

We performed sim2real transfer on a set of wing designs selected from the evolved NDF that spanned the range of {morphological simulation complexity} and predicted lift.   {Additionally, as an experimental control against our automated design process, we custom designed a limited number of wing morphologies by hand based on studies that have explored flapping wing aerodynamics.  The first design was a square ribbed plate with rigid ribs inspired by Sane \textit{et al} \cite{sane2002aerodynamic}. The second design is representative of the DelFly clapping wing construction \cite{de2009design} and included increased elasticity in the  wing structure to allow  deflection under aerodynamic loads, which is intuitively needed to create lift. Finally we created a minimal wing design representing the simplest design possible in the morphospace.  These wings were mapped into our paramaterised wing descriptor format for fair comparison in sim2real transfer.  }

%\RSSFIXME{This represents an automatically designed set of individuals each with predicted high performance.  We use this approach to select wings for the sim2real transfer rather than randomly generating a set of wing design that sample the morphospace which are inherently likely to be low performance.  Our focus on high performing designs in simulation allows measurement of the shape of the sim2real gap for design of interest rather than for all possible designs.}

%it is unclear that human design is intuition for design is beacyse there are no examples of flapping wing aircraft that are as efficient as natural fliers and aerodynamics changes substantially with Reynolds number (defined by size and speed) resulting in a non-intuitive design space. Regardless, 

%% file: 4_results.tex
\section{Results }

The design process was undertaken and produced the three non-dominated fronts, one of which is shown in figure \ref{fig:teaser}(c). The color of each individual represents $C_{MS}(m)$ for that design and shows that larger values of $C_{MS}$ allow larger predicted lift.  The non-dominated front shown included more than 100 morphology designs, produced efficient lift between 10mN and 200mN  and represent a spread across a large portion of the complexity space.

%The evolutionary search process was executed twice as described in the methods to maximise mean lift while  minimising drive power and drive torque respectively and resulted in non-dominated fronts  shown in Figure \ref{fig:evoDesignFronts}. As well as showing the front, the individual solutions within the front are coloured to match their morphological distance as described by Equation \ref{eqn:morphDist_defn}.  Each of these two non-dominated fronts of approximately 100 morphology designs each, produced efficient lift between 0mN and 200mN  and represent a spread across a large portion of the morphological distance.  From inspection of the fronts, the general dominance of the yellow coloring in the optimisation that reduce power rather than torque suggests that there is a relationship between morphological distances and reduced power consumption although we make no explanation here as to why.  

% Trimmed as one of them is in figure 1 a
% \begin{figure}[h!]
%   \centering
%     \includegraphics[width=0.48\textwidth]{images/evoDesignFront_LT_LP_composite.png}
% %    \includegraphics[width=0.48\textwidth]{images/evoDesignFront_LP.png}
% %    \includegraphics[width=0.48\textwidth]{images/evoDesignFront_LT.png}
%   \caption{Evolution produced non-dominated fronts selected for best balance. (a) mean lift versus mean Power. (b) mean lift vs peak torque }
%   \label{fig:evoDesignFronts}
% \end{figure}

%evenly spread and limited by space.   \RSSFIXME{R3 WHY 13?  Diverse is not a reason. KR - 4 batches of 4 one had smaller wings - limited by bench space}

 \begin{table}[t]
 \centering
\setlength\tabcolsep{2.5pt}
        \begin{tabular}[t]{|c|c|c|c|c|c|c|c|}
        \hline
        Wing  & Design &  $C_{MS}$  & $B(m)$  & $S(m)$ &  $L_S$  & $L_R$ & $STR$\\
        Label &       &       &         &  (mm)  &  (g)    &  (g)  & \\
        \hline
        MIN & \begin{minipage}{2cm}\includegraphics[width=\textwidth]{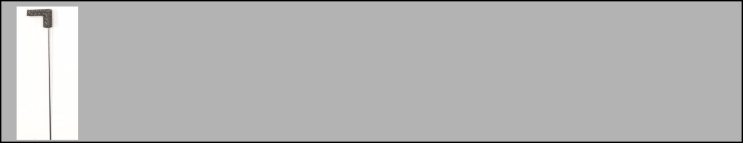}\end{minipage} & 0.13 & 1 & 50 & 0.0 & 0.0 $\pm$ 0.1 & 0.0 \\
        CD-A & \begin{minipage}{2cm}\includegraphics[width=\textwidth]{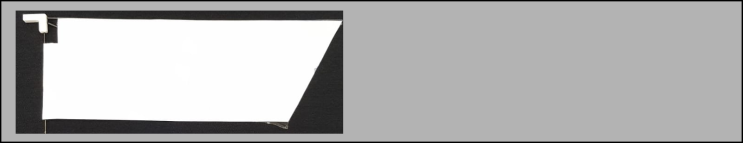}\end{minipage} & 0.4 & 2 & 250 & 0.0 & -0.5 $\pm$ 0.2 & -0.04 \\
        CD-B & \begin{minipage}{2cm}\includegraphics[width=\textwidth]{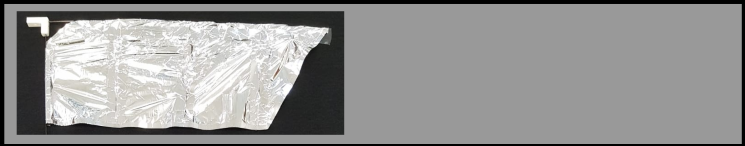}\end{minipage} & 0.49 & 3 & 250 & 3.8 & 6.4 $\pm$ 0.1 & 0.19 \\
        \hline
        EV-A & \begin{minipage}{2cm}\includegraphics[width=\textwidth]{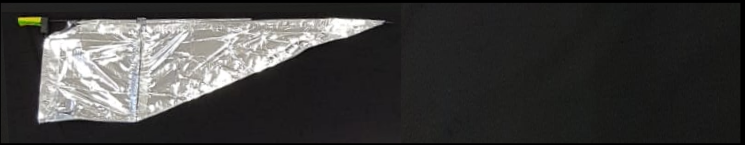}\end{minipage} & 0.46 & 2 & 308 & 3.4 & 7.3 $\pm$ 0.3 & 0.28 \\
        EV-B & \begin{minipage}{2cm}\includegraphics[width=\textwidth]{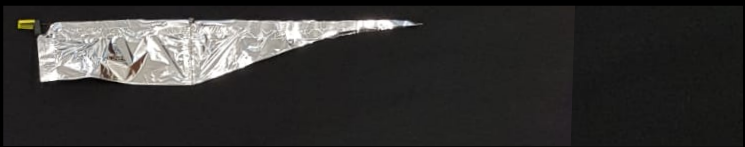}\end{minipage} & 0.5 & 2 & 352 & 3.8 & 5.0 $\pm$ 0.4 & 0.09 \\
        EV-C & \begin{minipage}{2cm}\includegraphics[width=\textwidth]{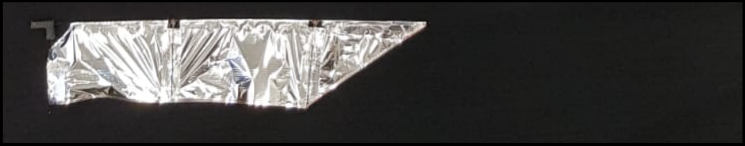}\end{minipage} & 0.54 & 3 & 308 & 2.9 & 10.1 $\pm$ 0.7 & 0.52 \\
        EV-D & \begin{minipage}{2cm}\includegraphics[width=\textwidth]{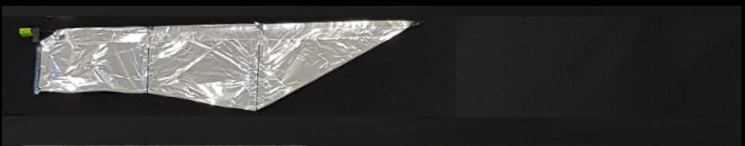}\end{minipage} & 0.57 & 3 & 341 & 4.6 & 6.0 $\pm$ 1.0 & 0.1 \\
        EV-E & \begin{minipage}{2cm}\includegraphics[width=\textwidth]{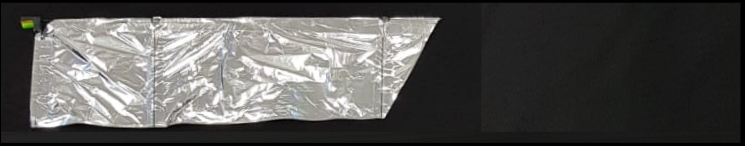}\end{minipage} & 0.6 & 3 & 365 & 7.0 & 6.3 $\pm$ 0.1 & -0.05 \\
        EV-F & \begin{minipage}{2cm}\includegraphics[width=\textwidth]{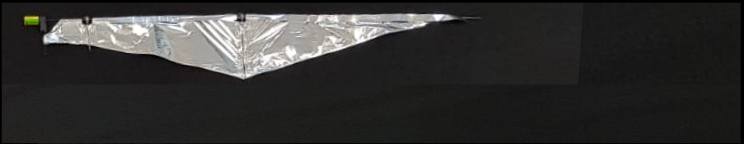}\end{minipage} & 0.61 & 3 & 383 & 5.2 & 6.2 $\pm$ 4.8 & 0.07 \\
        EV-G & \begin{minipage}{2cm}\includegraphics[width=\textwidth]{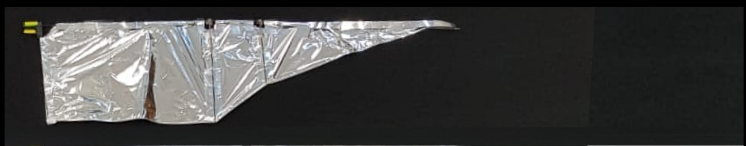}\end{minipage} & 0.62 & 3 & 393 & 8.0 & 7.2 $\pm$ 0.2 & -0.06 \\
        EV-H & \begin{minipage}{2cm}\includegraphics[width=\textwidth]{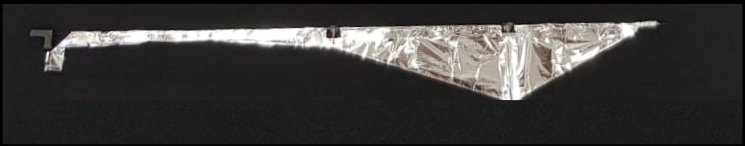}\end{minipage} & 0.71 & 3 & 479 & 12.3 & 2.3 $\pm$ 0.4 & -0.72 \\
        EV-I & \begin{minipage}{2cm}\includegraphics[width=\textwidth]{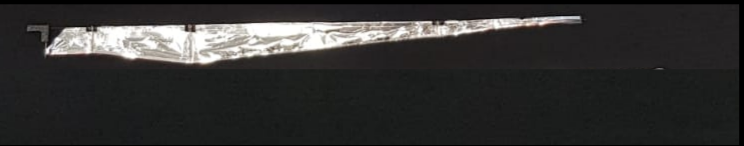}\end{minipage} & 0.75 & 4 & 443 & 5.4 & 2.4 $\pm$ 0.5 & -0.22 \\
        EV-J & \begin{minipage}{2cm}\includegraphics[width=\textwidth]{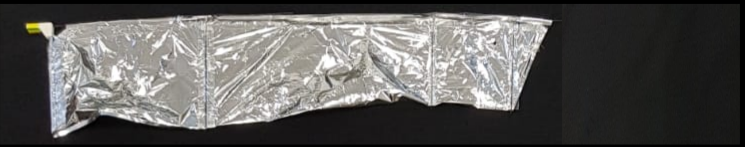}\end{minipage} & 0.76 & 4 & 454 & 13.0 & 4.5 $\pm$ 0.2 & -0.61 \\
        EV-K & \begin{minipage}{2cm}\includegraphics[width=\textwidth]{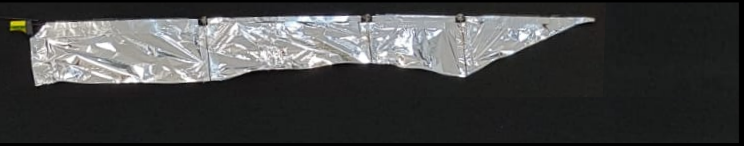}\end{minipage} & 0.82 & 4 & 517 & 13.1 & 3.2 $\pm$ 0.1 & -0.71 \\
        EV-L & \begin{minipage}{2cm}\includegraphics[width=\textwidth]{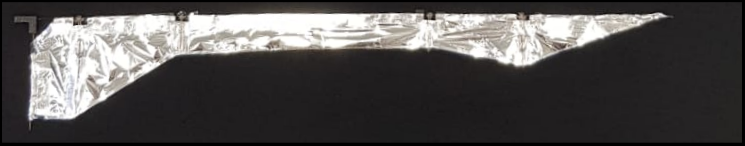}\end{minipage} & 0.9 & 5 & 509 & 10.1 & 1.8 $\pm$ 0.1 & -0.6 \\
        EV-M & \begin{minipage}{2cm}\includegraphics[width=\textwidth]{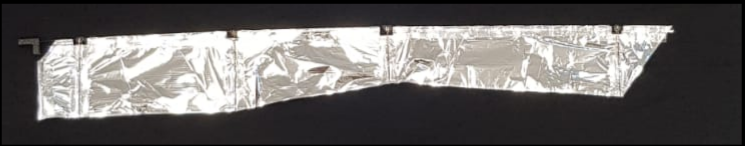}\end{minipage} & 0.92 & 5 & 526 & 13.9 & 4.4 $\pm$ 0.1 & -0.68 \\
         \hline
         \end{tabular} 
        \caption{Transferred wing designs and metrics}
        \label{tab:wingData}
        \vspace{-20pt}
  \end{table}

\begin{figure}
  \centering
    \includegraphics[width=\columnwidth]{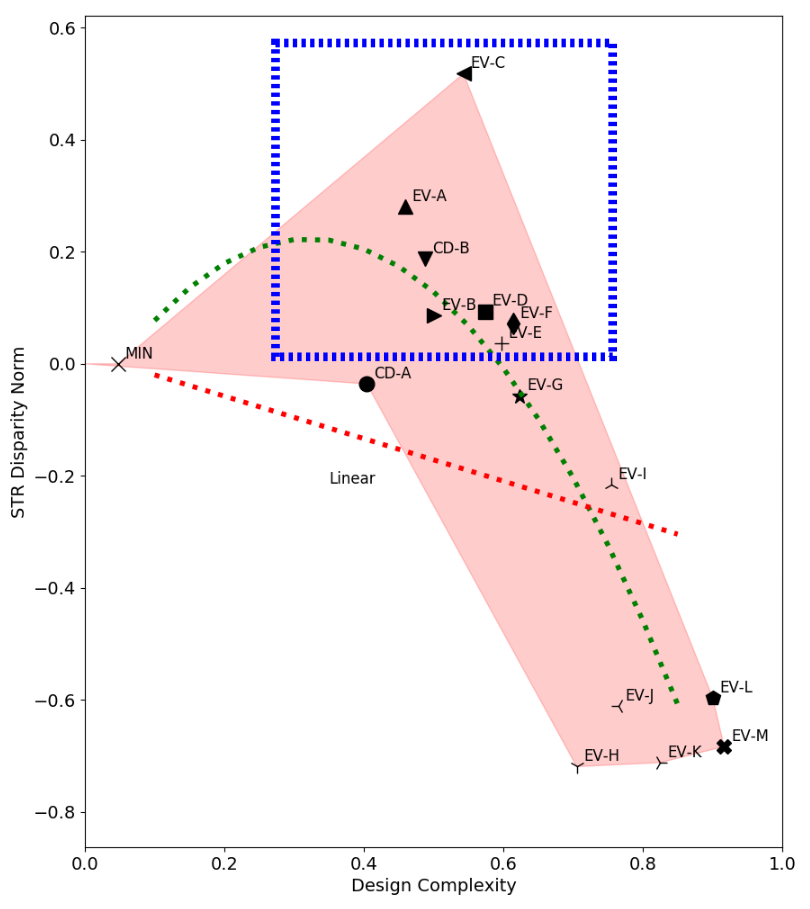}
  \caption{Scatter plot of $STR$ verses $C_{MS}$ for sim2real transfers.  Green dotted line - Polynomial best fit shape. Shaded region - bound of upper and lower $STR$ values of neighbouring sim2real transfers. {Red dotted line - monotonic decay prediction for $STR$ drawn for comparison. Blue bounded region - region where simulation under represents performance in reality. }}
  \label{fig:results_RGshape}
\end{figure}

Figure \ref{fig:results_RGshape} shows our key result from this study which is the scatter plot of $STR(m)$ against $C_{MS}(m)$ of the transferred wings.  That plot includes a bounded region (shaded in pink) set by connecting the upper and lower bound of $STR$ between adjacent transferred wings in the evolved search space.  Overlaid on the plot is the regression selected polynomial line of best fit ({second order}) for the transferred points (green dashed line).  The shape of that region and line of best fit represents the empirically measured shape of the reality gap for the flapping wing design.  

% Tried to clarify it as a boundary case - \RSSFIXME{R3: ZONE ARBITRARY? II and III could be one region?
% change conservative for "investment" - \RSSFIXME{R3: CONSERVATIVE means what here?}

% Added a summary in the intro and recall it here.  Enough?. \RSSFIXME{R1 + R3 - WHY INTUITIVE LINEAR - CLARIFY THIS}

{In our discussion of the reality gap shape from the introduction, we qualitatively predicted that $STR$ would degrade monotonically with $C_{MS}$, and we plot a best fit linear decay relationship extending from the minimal solution (red dotted line) in Figure \ref{fig:results_RGshape}.  It is clear from inspection that the empirically measured shaded region and polynomial best fit that true shape is substantially different.}  The shape of the reality gap that we have measured identifies a non-linear relationship, where $STR$ is small (less than $\pm0.2$) up to a threshold $C_{MS}$ value of approximately 0.6, at which point $STR$ falls off at a much faster rate. We interpret the rapid degradation in $STR$ for designs above the $C_{MS}$ threshold as a compounding of the smaller errors within the finite elements of our morphology simulation.

Further, in this example, we identify a region of the plot where substantially improved performance was obtained in reality compared to simulation.  This region is highlighted in blue, and represents an area where the model is insufficient to capture a crucial positive operating effect, which we interpret to be the unmodelled wake capture. Investment in simulation improvements in this zone are likely to result in improved search results, and development of methods to automatically improve the modelling environment in this zone would extend the automatic design capability.

Regardless of $STR$, Table \ref{tab:wingData} shows that the wing with the best real lift is ``EV-C" followed by ``EV-A" and ``EV-G" all of which outperformed the best hand designed wing, ``CD-B".  These wings all have lower aspect ratio than many of the other wings suggesting that there may be a contribution of that morphological similarity that directly affects lift. That same morphology group also have positive $STR$ and are exploiting one or more effects that do not appear in our simulator. Conversely, wings with higher aspect ratio tended to have negative $STR$ values and lower lift which suggests that this type of morphology leads to over-fitting to inaccuracies of the simulation that are not available to realised systems. In the future, these types of insights may be inferred automatically from transfer results and used to guide automatic finite element model improvement.

%% file: 5_discussion.tex
\section{Discussion }

For the first time in this study we have measured the shape of the reality gap that emerges during morphology-in-the-loop robotics design. In this instance, we found the novel result that the shape is non-monotonic. Up to a threshold complexity level, real behaviour of the robot matches or even exceeds simulation performance. Above that threshold, $STR$ quickly reduces resulting in the real performance of the robot being much worse that predicted.  
This study has produced one of only a few examples of morphological evolutionary search across the sim2real gap \cite{lipson2000automatic, cellucci20171d, mehta2014cogeneration,  coros2013computational}.   

Morphology is important for many robotics applications and there are many domains where human intuition may not be available for design.  Rather than using bio-mimicry, we have shown that a machine learning design process can create a diverse set of morphologies tailored for a task from an underlying set of bio-inspired characteristics.  Our approach led us to define a genotype that described a wing using features of shape, span, elasticity and inertia for both simulation and reality implementations of the flapping wing aerodynamics problem. We defined {morphological simulation complexity} based on measures relevant to the expected accuracy of finite element simulation. Using an evolutionary design process, we searched for high performance designs and showed that predicted lift in simulation increased with $C_{MS}$, but that real lift peaked at a threshold value of $\approx 0.6$ and degraded $STR$ and lift was found in the more complex designs.  We interpret this rapid degradation as the compounding rather than monotonic additive effect of smaller errors in the finite element morphology simulation structure.

We hypothesised that CFD would not be suitable for sim2real design due to computational overheads limiting search in simulation, and instead developed a fast, comparatively low fidelity simulation to facilitate automated search.  We found this revised approach was able to design high performance wings for successful sim2real transfer up to a threshold level of $C_{MS}$, and that those designs out-performed hand designed wings. This suggests that our simulation approach retains a gradient that evolution can follow towards high performing designs in reality.  Whether a similar reality gap shape would be found for a different simulation approach like CFD is a topic for future research.

{The results are encouraging for the utility of machine learning based sim2real morphology design using finite element simulation.  Had the shape of the gap been found to degrade proportionally (or worse) with design complexity, then an automated design target that produced maximum lift with minimum complexity would be the obvious search objective, but could only produce near trivial designs. For our problem, we have shown that a richer search space of wing morphologies that narrow the sim2real gap is available for exploration automatically for specific tasks.}

The reality gap shape for morphology-in-the-loop design we observed in this study is the first such measurement reported and it remains to be seen if this shape holds for other morphology domains. Our finite element simulation approach was applied linearly along the span of a flat wing, and an  interesting question would be to determine how dimensionality of the finite element structure would affect the relationship.  Applying a similar analysis to a 3 dimensional design problem, such as those in \cite{lipson2000automatic}, would assess how compounding of inaccuracies affects the sim2real relationship.

We focused this study on morphology effect on sim2real transfer and used a fixed control policy throughout. It remains to be seen how the sim2real gap would be affected when controllers are also included in to the search space.  For instance, are the morphologies we found in this study better able to cross the reality gap and produce lift than wings designed using different methods, or at random, when combined with tailored controllers?  %Evaluations with differing controllers would excite different frequency spectra which would likely affect sim2real transfer also.

%\RSSFIXME{The "transferability approach" \cite{koos2013transferability} for sim2real control design excludes the design region surrounding low measured transferability solutions such that it can automatically reach designs that transfer well.  The non-linear relationship of $STR$ in this study suggests that an exclusion region size for morphology design should be tuned based on $C_{MS}$ to avoid excluding nearby high quality solutions while also quickly excluding poor regions of the morphospace. }

Robust automatic robot design through the inclusion of sim2real feedback with our morphology exploration process is a topic for continued research and we identify two options for development. 
First, the transferability approach \cite{koos2013transferability} achieved controller design by taking feedback from sim2real transfers and excluding design space regions surrounding poor transfers.  We will investigate extension to morphology design by  incorporating the reality gap shape to tune region sizes based on $C_{MS}$.
Secondly, we identified in this application a region where performance in reality was better than in simulation. While we believe that this is due to the non-modeled wake capture effect and that directly incorporating this complex feature within the simulation would apply correction, our desire for an automatic design process suggests that investment in automated sim2real feedback to produce modelling updates of the finite element morphology simulation might reduce the need for human intuition in the simulation development. Using this, we hope to demonstrate the ability of biological evolution to build upon fortuitous morphological design features that improve performance without the requirement of human intuitive understanding.

Our intent is to take automated wing design from this one degree of freedom example to include multi-wing, multi-degree of freedom and variable control policies similar to the properties observed in natural fliers.  The scalability of our initial approach is suited to this.  Our test rig design in this study uses small sensors, low cost actuators and quickly constructed wings, and will allow us move from evaluating forces on a single tethered wing to multi-wing morphologies and to use varying control policies. Using this revision, we aim to undertake morphology and control selection to achieve lift as well as the force and torque vectoring demonstrated in stabilised hover and flight of biological fliers.